\pdfoutput=1
\documentclass[10pt,twocolumn,letterpaper]{article}
\usepackage[pagenumbers]{cvpr}

\ifdefined\pdfobjcompresslevel
\fi
\usepackage{tabularx,multirow,array}
\usepackage{microtype}

\newcommand{\methodtitle}{SkillHarness: Execution-Aware Coordination with Plug-and-Play Skills for Long-Horizon Traversal}

\definecolor{cvprblue}{rgb}{0.21,0.49,0.74}
\usepackage[breaklinks,colorlinks,allcolors=cvprblue]{hyperref}
\hypersetup{
  pdftitle={SkillHarness: Execution-Aware Coordination with Plug-and-Play Skills for Long-Horizon Traversal},
  pdfauthor={Bailin He, Xiting Zhao, Kang Wang, Xinyu Zhou, Yongxing Cui, Haojie Liu, Qiao Sun, Hongyan Zhang, Wenqiang Zhang},
  pdfsubject={Robotics and locomotion skill coordination}
}

\makeatletter
\patchcmd{\@maketitle}{\vspace*{24pt}}{\vspace*{12pt}}{}{}
\makeatother

\title{\methodtitle}
\author{\textbf{Bailin HE}\textsuperscript{1}, \textbf{Xiting Zhao}\textsuperscript{1,}\thanks{Corresponding authors: Xiting Zhao and Xinyu Zhou.}, \textbf{Kang Wang}\textsuperscript{2}, \textbf{Xinyu Zhou}\textsuperscript{3,\textdagger}, \textbf{Yongxing CUI}\textsuperscript{2}\\
  \textbf{Haojie Liu}\textsuperscript{2}, \textbf{Qiao Sun}\textsuperscript{2}, \textbf{Hongyan Zhang}\textsuperscript{4}, \textbf{Wenqiang Zhang}\textsuperscript{1}\\
  \textsuperscript{1}Fudan University, \textsuperscript{2}Lenovo Group Limited\\
  \textsuperscript{3}Nanyang Technological University, \textsuperscript{4}Shanghai Jiaotong University\\
  {\small\href{mailto:blhe25@m.fudan.edu.cn}{blhe25@m.fudan.edu.cn}, \href{mailto:zhaoxt@fudan.edu.cn}{zhaoxt@fudan.edu.cn}, \href{mailto:xinyu.zhou@ntu.edu.sg}{xinyu.zhou@ntu.edu.sg}}
}
\date{}

\begin{document}
\setcounter{footnote}{1}
\maketitle
\begin{abstract}
Traversing complex terrain requires robots to coordinate diverse locomotion skills while adapting to unseen obstacles and changing environments. Although perceptive reinforcement learning enables agile locomotion, conventional multi-terrain locomotion rely on fixed policy, limiting adaptability and skill extensibility. Perceptual disturbances and unreliable skill transitions further challenge robust execution. To address these challenges, we propose SkillHarness, an execution-aware coordination framework that integrates a flexible, plug-and-play locomotion skill library. To enhance traversal robustness, we introduce a perceptive reinforcement learning approach that preserves task-relevant environmental references while continuously updating proprioceptive feedback, reducing sensitivity to perceptual fluctuations. A vision-language-guided Harness coordinates navigation and skill execution through event-triggered reasoning, geometric feasibility verification, and execution-phase constraints. By flexibly composing and extending available skills, SkillHarness adapts to unseen terrain and task combinations without retraining the locomotion policies. The framework further incorporates execution feedback and retrospective decision revision to improve skill selection and avoid inappropriate skill switching. We evaluate SkillHarness across diverse terrain configurations, perceptual disturbances, and multi-obstacle traversal tasks to assess its adaptability and reliability in long-horizon traversal.

\end{abstract}

\section{Introduction}
\label{sec:intro}

Complex terrain traversal is a fundamental capability for autonomous
mobile robots in unstructured environments, supporting industrial
inspection, disaster response, and outdoor exploration. Unlike navigation
on relatively flat ground, it requires coordinating skills such as
walking over uneven surfaces, crouching through confined spaces,
crossing gaps, and climbing platforms. 
Reliable long-horizon traversal requires not only successful execution of individual locomotion skills but also coordinated navigation, appropriate skill selection, and continuous progress toward the task objective. 
Unseen obstacle combinations, environmental changes, and perceptual disturbances can
lead to incorrect skill selection or inappropriate switching,
interrupting subsequent execution. Individual skill performance and
complete-task reliability must therefore be considered together.

Perceptive reinforcement learning (RL) has significantly advanced robot mobility over challenging terrain, including methods based on learned terrain perception and robust state estimation~\cite{he2026surfloco,long2025pim,dong2026mgdp,liu2026real}. Multi-skill approaches further extend locomotion capabilities. ANYmal Parkour learns a high-level navigation policy to coordinate specialized skills~\cite{hoeller2024anymalparkour}, while Parkour in the Wild distills terrain experts into a unified controller and expands its capabilities through RL fine-tuning~\cite{rudin2026parkourwild}. However, adapting learned coordinators or unified controllers to new skills requires additional training, introducing computational costs and potential interference with previously learned behaviors.

Recent advances in language-guided robotics demonstrate the
potential of semantic reasoning for organizing reusable robotic
capabilities. SayCan grounds language plans in learned skill
affordances~\cite{ichter2023saycan}, while NaVILA connects
vision-language reasoning with legged locomotion through
spatial commands~\cite{cheng2025navila}.
More recently, Harness VLA combines memory-guided planning,
analytic primitives, and a frozen VLA to improve long-horizon
manipulation without fine-tuning~\cite{zhang2026harnessvla}.
These advances motivate exploring how an execution-aware
Harness can organize flexible locomotion skills for
long-horizon terrain traversal.

Beyond semantic coordination, reliable traversal also
requires robust execution of individual locomotion skills.
High-impact contacts and occlusion can degrade perception
and state estimation~\cite{hoeller2024anymalparkour}.
Human obstacle-crossing studies suggest that visual information
acquired during approach can support subsequent maneuvers,
with different requirements for online vision across
movement stages~\cite{timmis2012obstacle,diaz2018pickup}.
This motivates phase-dependent management of environmental
references during critical locomotion maneuvers.

To this end, we propose SkillHarness, a plug-and-play
locomotion framework built around a unified Execution-Aware
Harness. Each skill exposes a common interface specifying
its semantic role, capability limits, observation adapter,
and execution constraints, allowing skills to be added,
removed, or replaced without retraining existing
locomotion policies or vision-language models.
A lightweight Gemma-based trigger determines when semantic
reasoning is needed, while asynchronous DeepSeek-based
reasoning generates local navigation goals and candidate skills.

Before execution, the Harness validates asynchronous proposals
against current task states, obstacle geometry, and
execution-phase constraints. Navigation goals and locomotion
skills are updated only when the corresponding conditions are
satisfied, while critical maneuvers are protected from
inappropriate switching. Execution feedback monitors skill
initiation, completion, and failure to support subsequent
decision revision. During critical phases, skill-specific
adapters retain relevant environmental references while
continuously updating proprioceptive feedback and action history.

We evaluate SkillHarness across structured and unstructured
terrain combinations using multi-obstacle traversal routes,
comparisons with perceptive locomotion baselines, and Harness
verification ablations in simulation and on the physical robot.
Skill-level experiments further examine traversal with retained
environmental references. Offline execution logs support
retrospective refinement of decision rules and prompts without
updating locomotion policies.

The main contributions are as follows:
\begin{itemize}

    \item \textbf{A plug-and-play locomotion skill library} enables flexible skill addition, removal, and replacement through unified interfaces, without retraining existing locomotion policies or vision-language models.

    \item \textbf{A dual-model Execution-Aware Harness} combines lightweight Gemma-based event triggering with asynchronous DeepSeek-based reasoning to coordinate navigation and skill execution, enabling flexible adaptation to diverse terrain and task combinations.
    
    \item \textbf{A robust closed-loop execution mechanism} integrates geometric verification, execution-phase constraints, and feedback-driven decision revision to prevent inappropriate skill switching, while preserving task-relevant perceptual references for reliable traversal.
\end{itemize}

\section{Related Work}
\label{sec:related}

\paragraph{Perception for legged locomotion.}
The Perceptive Internal Model (PIM) uses continuously updated elevation maps for humanoid locomotion~\cite{long2025pim}. MGDP combines depth denoising with terrain-adaptive learning for quadruped control~\cite{dong2026mgdp}, and REAL integrates temporal visual memory and physics-guided state estimation~\cite{liu2026real}. PUMA learns egocentric foothold priors~\cite{wang2026puma}, while StairMaster combines depth-noise modeling with spatiotemporal memory for hollow-stair traversal~\cite{tang2026stairmaster}.
RPL distills terrain-specific experts into a perceptive policy using multi-view depth observations~\cite{zhang2026rpl}. 
However, maintaining high traversal success rates across unstructured terrain remains challenging, particularly under varying obstacle geometries and perceptual disturbances.

\paragraph{Multi-skill locomotion and extensibility.}
ANYmal Parkour trains a high-level navigation policy to coordinate specialized locomotion policies~\cite{hoeller2024anymalparkour},
and Yu et al. learn state-dependent switching criteria for agile
motions~\cite{yu2026switching}. Parkour in the Wild distills terrain-specific experts into a unified controller and extends its capabilities through additional RL fine-tuning~\cite{rudin2026parkourwild}.
APT-RL uses trajectory-optimized motion priors to train a unified multi-skill policy~\cite{kang2026apt}, and Perceptive Humanoid Parkour combines motion matching with expert distillation~\cite{wu2026php}.
However, changing real-world terrain and task requirements motivate skill architectures that allow individual locomotion policies to be added, updated, or replaced without time-costly training or optimization.

\paragraph{Vision-language planning and coordination.}
SayCan grounds language plans in learned skill affordances~\cite{ichter2023saycan},
while NaVILA generates spatial language commands for legged navigation~\cite{cheng2025navila}.
WayTrace translates map directions into image-space traces
for RL navigation~\cite{patel2026waytrace}. AsyncVLA separates remote semantic reasoning from onboard
reactive control~\cite{hirose2026asyncvla}, while Xu et al.
combine asynchronous perception with spatial-semantic memory
for navigation~\cite{xu2026hierarchicalvln}.
Building on these advances, extending vision-language reasoning to jointly coordinate navigation and an evolving repertoire of terrain-specific locomotion skills remains a promising research direction.

\paragraph{Execution monitoring and correction.}
Harness VLA uses memory-guided planning and analytic primitives to stage and retry a frozen manipulation VLA~\cite{zhang2026harnessvla}, while VLA-Corrector uses a stage-aware verifier to detect failures and trigger prompt-based recovery~\cite{song2026vlacorrector}.
Extending these ideas to dynamic locomotion requires validating delayed proposals against current terrain geometry and execution phases, while monitoring skill completion and failure.

\section{Method}
\label{sec:method}
\begin{figure*}[t]
\centering
\includegraphics[width=\textwidth]{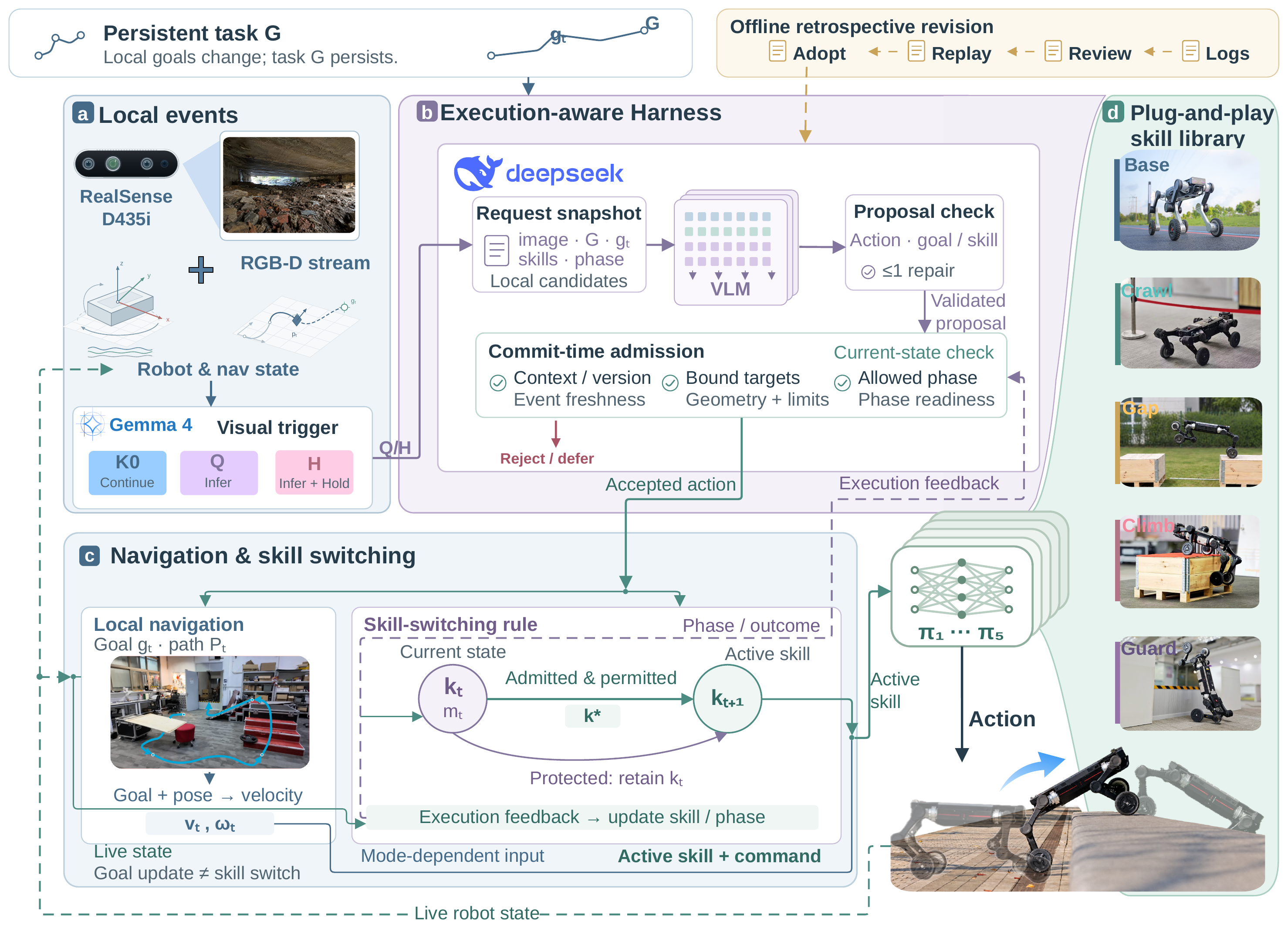}
\caption{\textbf{SkillHarness system overview.}
(a) RGB-D observations and robot/navigation state drive a local
Gemma~4-E2B visual trigger that continues execution, requests reasoning, or additionally
requests a controlled hold.
(b) The Harness constructs a request snapshot and local candidates,
checks the DeepSeek~V4.1~Flash proposal, and performs admission checks against the
current execution state.
(c) Accepted actions update navigation or request a phase-permitted
skill switch; execution feedback updates skill and phase state.
(d) Independently trained skills receive inputs through their own
adapters. The task objective $G$ persists as local goals change.
The offline path reviews and replays logs before adopting revisions.}
\label{fig:framework}
\end{figure*}

SkillHarness coordinates independently trained locomotion skills around
a persistent task objective. Figure~\ref{fig:framework} organizes the
system into local event detection, an execution-aware Harness,
navigation and skill switching, and a plug-and-play skill library.
The Harness rechecks asynchronous semantic proposals against the live
execution state before admitting commands. During a maneuver, phase
permissions constrain command changes, while skill-specific adapters
control which perceptual inputs remain live or are retained. Execution
feedback closes the online loop, and logged failures support offline
revision of the decision configuration.

\subsection{Task State and Skill Interfaces}
\label{sec:skill_interface}
Let $G$ be the persistent task objective and $g_t$ the current local
goal. The execution state $s_t$ includes robot feedback, navigation
state, the active skill $k_t$, its phase $m_t$, and task and plan
revisions. Local decisions can update $g_t$ while preserving $G$.
The library in Fig.~\ref{fig:framework}(d) contains Base, Crawl, Gap,
Climb, and Guard. Base follows navigation commands between obstacles.
Climb denotes ascent onto a single raised platform.

Each skill supplies a contract comprising its semantic role, parameter
bounds, observation adapter, phase permissions, and completion
conditions. The active policy produces
\begin{equation}
 \mathbf u_t=\pi_{k_t}\!\left(
 \Phi_{k_t}(\mathbf x_t,\mathbf c_t,R_t,\mathbf h_t)\right),
 \label{eq:adapter}
\end{equation}
where $\mathbf x_t$ is robot feedback, $\mathbf c_t$ is the
mode-dependent command, $R_t$ is the obstacle reference when required,
and $\mathbf h_t$ is skill history. The adapter $\Phi_k$ selects and
transforms the navigation commands, structural measurements, and
feedback required by $\pi_k$. Observation dimensions, history lengths,
and retained channels can differ across skills. Integrating a skill
requires its adapter and contract while keeping the other policy
weights fixed.

\subsection{Local Events and Asynchronous Proposals}
\label{sec:trigger}
In Fig.~\ref{fig:framework}(a), RGB-D sensing supplies the current image
$I_t$ and local geometry. A visual trigger reads $I_t$ and a compact
robot/navigation state summary $\xi_t$, and emits
\begin{equation}
 \eta_t=\mathcal T(I_t,\xi_t),\quad
 \eta_t\in\{\mathrm{K0},\mathrm{Q1{:}4},\mathrm{H1{:}4}\}.
\end{equation}
K0 continues execution. Q events request reasoning while the active
controller can continue moving. H events also request a controlled
hold, which takes effect only when the current phase permits it.
The four event reasons are obstacle, target, replanning, and
uncertainty. In our implementation, Gemma~4-E2B serves as the local
visual trigger, while DeepSeek~V4.1~Flash generates asynchronous high-level
proposals. The trigger determines when to request intervention, and
the VLM reasoner in Fig.~\ref{fig:framework}(b) proposes a local action.

For request $r$, the Harness captures a snapshot $S^{(r)}$ containing
the task, local goal, scene, available skills, and execution phase,
along with locally generated candidates $\mathcal V^{(r)}$.
The VLM asynchronously produces a structured proposal
\begin{equation}
 \rho^{(r)}=\mathcal F(I^{(r)},G,S^{(r)},\mathcal V^{(r)}).
\end{equation}
Its action specifies whether to continue, navigate, plan a traversal,
execute a skill, adjust posture, complete the task, or request
information. A navigation proposal selects a supplied local goal by
identifier; a skill proposal binds a skill to an obstacle reference.
Planning records intent before execution. The proposal check validates
the action schema and permits at most one repair after a schema
failure. A schema-valid response proceeds to admission checking;
it does not itself authorize execution.

\subsection{Geometry and Commit-Time Admission}
\label{sec:admission}
Local perception supplies obstacle boundaries, support surfaces, and
clearances, with reference identities and validity information.
A goal candidate contains its request-time coordinates $g_i^{(r)}$
and path $P_i^{(r)}$. The Harness tests this original pair against the
current state through
$C_{\rm goal}=\operatorname{valid}(g_i^{(r)},P_i^{(r)},s_t)$.
Changing either the coordinates or the path requires a new candidate
binding.

For a world-frame boundary with unit normal $\mathbf n$ and offset
$b$, body position $\mathbf r_t$ and yaw $\psi_t$ give
\begin{equation}
\begin{aligned}
 d_t&=b-\mathbf n^{\mathsf T}\mathbf r_t,\\
 e_{\psi,t}&=\operatorname{wrap}
 (\operatorname{atan2}(n_y,n_x)-\psi_t).
\end{aligned}
\label{eq:geometry}
\end{equation}
Guard maps this relative geometry into its training coordinates.
Gap constructs its reference frame from the near edge, measured
width, boundary normal, and support height. These metric inputs enter
the corresponding skill adapter.

Before committing a proposal, the Harness checks its locally bound
context against the live execution state,
\begin{equation}
\begin{aligned}
 A(\rho^{(r)},S^{(r)},s_t)
 ={}&C_{\rm context}\land C_{\rm event}\land C_{\rm age}\\
 &\land C_{\rm revision}\land C_{\rm executable}.
\end{aligned}
\label{eq:admission}
\end{equation}
These checks correspond to context/version, event freshness, bound
targets, geometry/limits, and phase readiness in
Fig.~\ref{fig:framework}(b). Context binds the session, task, episode,
source, and scene. The event must remain active, the response must
remain within its validity interval, and plan, hold, skill, and phase
revisions must match. Executability covers fault state, capability
bounds, candidate and goal readiness, current structural evidence,
and permission to update goals or switch skills.

A gap proposal can therefore become inadmissible during inference if
the robot enters a protected maneuver or its bound approach path
becomes untraversable. A failed check rejects or defers the proposal;
an admitted action is passed to navigation and skill switching.
Admission establishes software permission to issue a command.
Physical traversal still depends on contact, state estimation, and
motor response.

\subsection{Navigation and Phase-Constrained Execution}
\label{sec:execution}
Figure~\ref{fig:framework}(c) separates local navigation from skill
switching. Given the local goal $g_t$, path $P_t$, and current pose,
navigation produces a planar command
\begin{equation}
 \mathbf c_t^{\rm nav}
 =\left(v_{x,t},v_{y,t},\omega_{z,t}\right),
 \label{eq:navigation_command}
\end{equation}
supporting omnidirectional translation
$\mathbf v_t=(v_{x,t},v_{y,t})$ and yaw. Base consumes the
navigation command through its adapter; traversal skills receive the
mode-dependent inputs specified by their contracts. Updating $g_t$
changes the navigation target. A skill switch requires an admitted
execution request and permission from the current phase.

Let $k^*$ be the requested skill and let $\mathcal P(k_t,m_t,k^*)$
denote the phase permission for this switch. For a pending skill
request, the switching rule is
\begin{equation}
 k_{t+1}=\begin{cases}
 k^*, & A=1\ \land\ \mathcal P(k_t,m_t,k^*)=1,\\
 k_t, & \text{otherwise}.
 \end{cases}
 \label{eq:switch}
\end{equation}
Guard protects approach and maneuver against holds and switches.
Gap protects execution until completion or failure. Climb restricts
changes during support transfer, and Crawl protects its lowering and
raising phases. Semantic inference can continue during these phases,
while the active skill is retained when a requested change is not
permitted. Completion and failure feedback also update the skill and
phase, including the return to Base.

Execution requests are correlated with their feedback. Repeated
requests retain the original measurement identity and age. Controller
events distinguish initiation, cancellation, return to Base, and Gap
completion. The Harness tracks permission, motor onset, and traversal
completion separately, allowing route navigation to resume after the
relevant transition. Feedback updates the live state used by both the
trigger and subsequent admission checks. Loss of execution authority
or required feedback revokes permission, and manual control retains
priority. Physical success is evaluated using the traversal and
terminal conditions in Sec.~\ref{sec:experiments}.

\subsection{Single-Reference Skill Execution}
\label{sec:single_reference}
Each skill in the library shown in Fig.~\ref{fig:framework}(d)
uses a dedicated observation adapter to manage perceptual inputs
during execution. The adapter determines which channels remain
live and which are retained during critical phases. Reference
acquisition and actor-input retention occur at different points
within a maneuver. During approach, a skill can update its obstacle
reference $R_t$ and compute relative features
$\widetilde{\mathbf z}_t=\varphi(R_t,\mathbf x_t)$.
Locking $R_t$ at $R^*$ fixes the obstacle reference, while relative
distance and heading can still change with body motion through
Eq.~\ref{eq:geometry}. A binary channel mask $M_k$ specifies which
actor inputs are retained for $\tau_k\leq t<\tau_{\rm release}$,
\begin{equation}
 \mathbf z_t=(1-M_k)\odot\widetilde{\mathbf z}_t+
 M_k\odot\widetilde{\mathbf z}_{\tau_k}.
 \label{eq:retention}
\end{equation}
Here $\tau_k$ is the capture time and $\tau_{\rm release}$ ends the
retention interval. This notation summarizes each adapter's channel
rule. Proprioception and action history continue to update.

Gap updates its reference during approach and captures the entire
perception vector
$[w,x_{\rm near}-x,\operatorname{wrap}(-\psi)]$ at takeoff.
Here $w$ is gap width, and $x_{\rm near}$, $x$, and $\psi$ are defined
in the skill frame. The captured width, distance, and heading remain
fixed during retention. The adapter clears its perception and skill
vectors at completion or failure. Guard retains selected
localization-pose channels; its relative distance and heading follow
their own update rules. These adapters therefore preserve different
inputs across their critical phases.

\paragraph{Offline retrospective revision.}
SkillHarness supports offline refinement of prompts and decision rules
through the logs--review--replay--adopt cycle in
Fig.~\ref{fig:framework}. Execution logs and failure cases are reviewed
to identify conditions requiring revision. Candidate updates are
replayed against recorded cases before adoption in subsequent runs.
This process updates the Harness configuration while keeping model
and motor-policy weights fixed. Online execution uses the adopted
configuration together with live-state admission checks, phase
constraints, and execution feedback.

\section{Experiments}
\label{sec:experiments}
We compare multi-terrain route completion with MGDP and REAL, evaluate
the effect of the Harness pre-execution check in simulation and on the
physical robot, and examine the skill-level execution and model choices
within SkillHarness.

\begin{figure*}[!t]
\centering
\includegraphics[width=\textwidth]{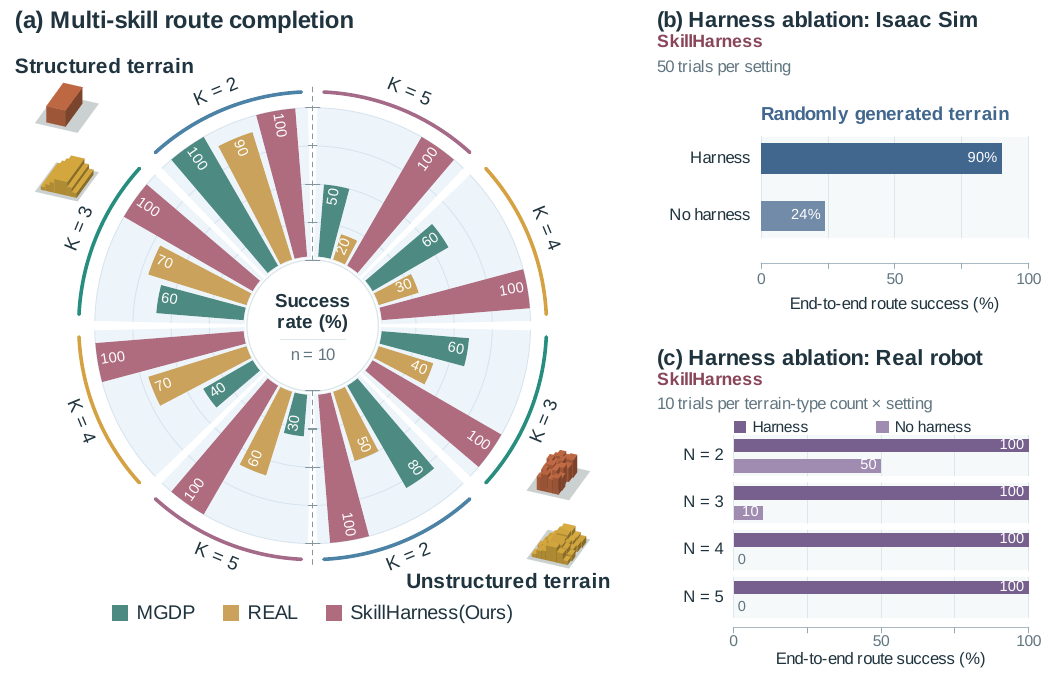}
\caption{\textbf{Multi-terrain route completion and Harness verification.}
(a) MGDP and REAL run in MX simulation, and SkillHarness runs on the
physical robot. $K$ denotes the number of distinct skill types required
by a route, with ten trials per method and route.
(b)~Harness ablation in Isaac Sim on randomly generated routes
containing five terrain segments, with repeated obstacle types allowed
and 50 trials per setting.
(c)~Physical-robot ablation with $N\in\{2,3,4,5\}$ distinct terrain
types per route and ten trials per terrain count and setting.
The Harness rule set was refined using the five failed Harness trials
in (b) before deployment in (c).
All panels report end-to-end route success.}
\label{fig:route_and_ablation}
\end{figure*}

\subsection{Evaluation Setting}
We compare MGDP~\cite{dong2026mgdp}, REAL~\cite{liu2026real},
and SkillHarness on multi-terrain route completion.
We evaluate Harness verification in Isaac Sim and on the physical robot. Table~\ref{tab:capabilities} lists
coverage for the five terrain types of interest and the command
interfaces used in these configurations. Stair denotes a flight of
steps and Climb denotes a single raised platform. Base provides
navigation between obstacles. SkillHarness exposes lateral velocity
in addition to forward velocity and yaw rate, allowing lateral
approach corrections in the navigation segments.

\begin{table}[!ht]
\centering\scriptsize
\setlength{\tabcolsep}{2.6pt}
\caption{\textbf{Obstacle and command coverage.} A check mark indicates
confirmed support in the evaluated configuration. A cross indicates
that the capability is outside its confirmed coverage. MGDP's
\emph{Narrow Crawl} corresponds to Crawl.}
\label{tab:capabilities}
\begin{tabular}{@{}lcccccc@{}}
\toprule
Method & Stairs & Crawl & \shortstack{Rodent\\barrier} & Gap &
\shortstack{High\\platform} & \shortstack{Motion\\commands}\\
\midrule
MGDP~\cite{dong2026mgdp} &
$\checkmark$ & $\checkmark$ & $\times$ & $\checkmark$ & $\times$ &
\shortstack{Forward + yaw\\$(v_x,\omega_z)$}\\
REAL~\cite{liu2026real} &
$\checkmark$ & $\times$ & $\times$ & $\checkmark$ & $\checkmark$ &
\shortstack{Forward + yaw\\$(v_x,\omega_z)$}\\
\addlinespace[1pt]
SkillHarness &
$\checkmark$ & $\checkmark$ & $\checkmark$ & $\checkmark$ & $\checkmark$ &
{\setlength{\fboxsep}{2pt}\colorbox{blue!8}{\begin{tabular}[c]{@{}c@{}}
  \textbf{Omnidirectional}\\[-1pt]
  \textbf{translation + yaw}\\[-1pt]
  $\boldsymbol{(v_x,v_y,\omega_z)}$
  \end{tabular}}}\\
\bottomrule
\end{tabular}
\end{table}

The baselines use the same D435i viewpoint in simulation.
SkillHarness uses its calibrated robot camera pose. Within each
method we fix the checkpoints, command interface, and applicable
planner configuration. For the route-completion comparison, we select
evaluation layouts separately from training and prompt development.
For each evaluated route condition, we fix the reset procedure,
start and goal, and time budget. The simulation ablation additionally
provides failure cases for refining the Harness rules before the
physical-robot ablation (Sec.~\ref{sec:harness_experiment}).

\subsection{Multi-Terrain Route Completion}
\label{sec:route_experiment}
We evaluate routes requiring $K\in\{2,3,4,5\}$ distinct skill types
under both structured and unstructured layouts. Each method runs
ten independently reset trials per route. In the MX baseline tests,
the routes contain $K$ obstacles and $K$ navigation targets, so
the obstacle and target counts increase with skill diversity.
The unstructured variants change local geometry or placement,
for example through varying stair risers, skewing a gap edge, or
offsetting a platform edge. SkillHarness receives the terminal
goal and selects skills online, resuming navigation between obstacles.
A trial succeeds if the robot autonomously clears all designated
obstacles in sequence, makes only permitted contacts, and stops
within the terminal tolerance before the deadline. All starts,
including interventions and timeouts, enter the denominator. Table~\ref{tab:skillharness_terrain} gives the
nominal SkillHarness capability ranges used for route design.
For Crawl, the parameter is the commanded body height.

\begin{table}[!ht]
\centering\small
\setlength{\tabcolsep}{4pt}
\caption{\textbf{SkillHarness terrain support.} Geometry and command
ranges for route design.}
\label{tab:skillharness_terrain}
\begin{tabular}{@{}llc@{}}
\toprule
Terrain & Parameter & Range (m)\\
\midrule
Stairs & Riser height & $\leq 0.23$\\
Crawl & Adjustable body height & 0.21--0.51\\
Rodent barrier & Barrier height & 0.40--0.70\\
Gap & Gap width & 0.30--1.60\\
High platform & Platform height & 0.30--1.00\\
\bottomrule
\end{tabular}
\end{table}

\paragraph{Comparison with MGDP and REAL.}
As $K$ increased from 2 to 5, success on the tested unstructured
routes fell from $80\%$ to $50\%$ for MGDP and from $50\%$ to $20\%$
for REAL. On structured routes, the corresponding rates fell from
$100\%$ to $30\%$ for MGDP and from $90\%$ to $60\%$ for REAL.
SkillHarness completed all ten trials in each of the eight route
conditions (Fig.~\ref{fig:route_and_ablation}(a)). At $K=5$,
SkillHarness exceeded the higher-scoring baseline by 40 percentage
points on the structured route (REAL, $60\%$) and by 50 points on
the unstructured route (MGDP, $50\%$).

In the baseline tests, the numbers of required skill types, obstacles,
and navigation targets increased jointly with $K$. Terrain
combinations and difficulty levels were randomly sampled, so
structured and unstructured routes with the same $K$ were not
necessarily matched in traversal difficulty. At a fixed $K$,
REAL scored 30--40 percentage points lower on unstructured routes,
whereas MGDP scored 20 points higher at $K=4$ and $K=5$.
This variation can be explained by the sampled terrain conditions:
an unstructured route may contain geometries and difficulty levels
better suited to MGDP's learned capabilities than those encountered
on a structured route. The observed differences therefore depend
on the sampled terrain combinations and difficulty levels, as well
as on whether the layout is structured or unstructured.

Overall, as the number of required skill types increased, MGDP
and REAL exhibited declining route-completion rates. Both methods
also showed limited reliability on the tested unstructured routes,
with performance varying across terrain combinations and difficulty
levels. SkillHarness maintained $100\%$ observed success across
all eight tested conditions, demonstrating reliable multi-skill
coordination across increasingly diverse terrain combinations
and unstructured layouts.

\subsection{Harness Verification Ablation}
\label{sec:harness_experiment}

We compare route completion with the Harness pre-execution check
in Eq.~\ref{eq:admission} enabled or disabled. Within each platform,
the two settings share the VLM, prompt, trigger conditions, candidates,
skill policies, low-level executor, and termination criteria.
Only the high-level verification of proposed actions is varied.

\paragraph{Isaac Sim.}
Figure~\ref{fig:route_and_ablation}(b) reports 50 trials per setting
on randomly generated routes containing five terrain segments,
with repeated obstacle types allowed.
Harness achieved an end-to-end success rate of $90\%$ ($45/50$),
compared with $24\%$ ($12/50$) without Harness, a gain of
66 percentage points.

\paragraph{Failure analysis and rule refinement.}
We evaluated the VLM on 100 obstacle photographs, asking it
to select the skill required to traverse the obstacle shown
in each image. The model selected the correct skill in 91
cases, yielding an image-level skill-selection accuracy of
$91\%$. However, high accuracy on individual images did not
ensure appropriate decisions during motion. For example,
a staircase visible behind a raised platform could prompt
a premature selection of Base, although the platform
immediately ahead still required Climb. Changes in the
camera view during locomotion could also perturb proposals
and lead to mistimed skill transitions.

The five failed Harness trials in simulation were subsequently
analyzed to identify missing rules for the encountered traversal
conditions. These cases informed additional rules concerning
obstacle order, task progress, and switching permissions during
motion. The $90\%$ success rate in panel (b) was measured before
these updates. The refined rule set was then used in the physical-robot
experiments in Fig.~\ref{fig:route_and_ablation}(c).
Thus, the simulation evaluation also provided failure cases for
improving rule coverage before physical deployment.

\paragraph{Physical robot.}
Let $N$ denote the number of distinct terrain types in a route.
We evaluated $N\in\{2,3,4,5\}$, with ten trials per terrain count
and setting. Using the refined Harness rules, the robot completed
all ten trials at every $N$. Without Harness, success was
$50\%$ ($5/10$) at $N=2$, $10\%$ ($1/10$) at $N=3$,
and $0\%$ at both $N=4$ and $N=5$.

Without Harness, the robot could still complete some simpler
routes containing terrain well handled by its existing motor
skills. However, this capability did not ensure reliable
coordination across different terrain types: when five distinct
types were combined into a single route, none of the ten trials
succeeded. These results highlight the importance of selecting
the skill required by the immediate obstacle and admitting the
switch at an appropriate execution phase.

Overall, the results support the role of Harness in maintaining
long-horizon navigation and coordinating accurate, timely skill
transitions. By combining the persistent task objective with
current terrain geometry, execution state, and accumulated
traversal rules, Harness helps the robot complete successive
obstacles and resume navigation toward the final goal.

\subsection{Skill-Level and Model Evaluation}
\label{sec:reference_trials}

\begin{figure}[!htbp]
\centering
\includegraphics[width=\linewidth]{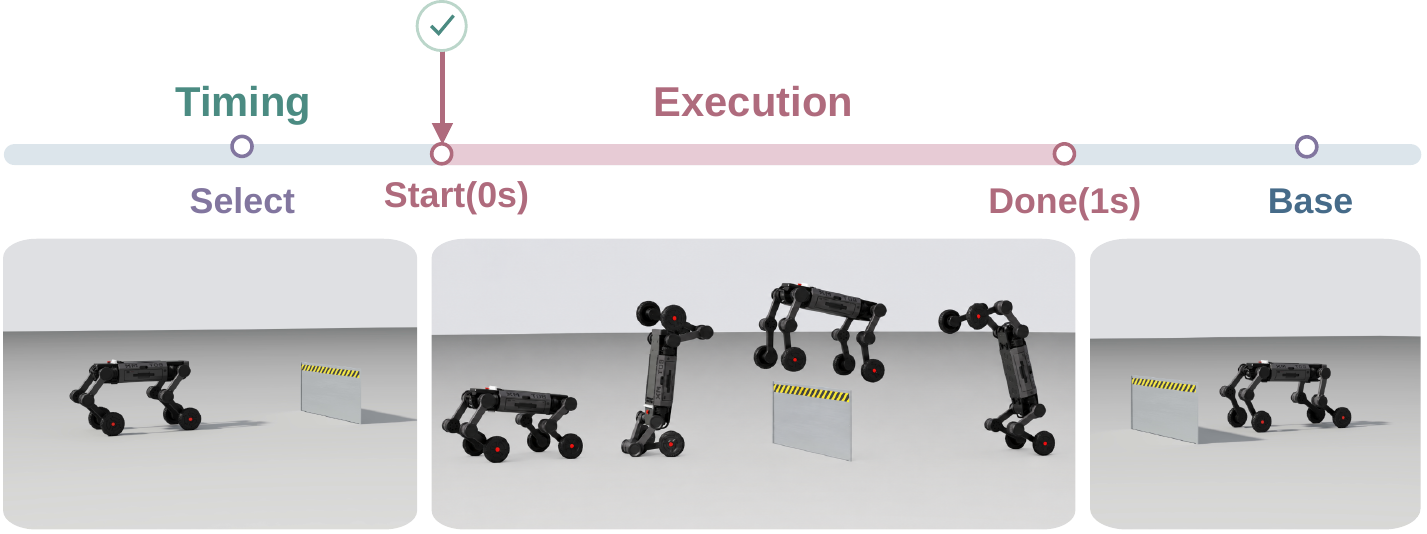}
\caption{\textbf{Guard maneuver illustrated in simulation.}
The sequence shows approach, rodent-barrier crossing, and exit.
The reported success counts are obtained from separate
physical-robot trials.}
\label{fig:guard_sequence}
\label{fig:qualitative}
\end{figure}

\begin{figure}[!htbp]
\centering
\includegraphics[width=\linewidth]{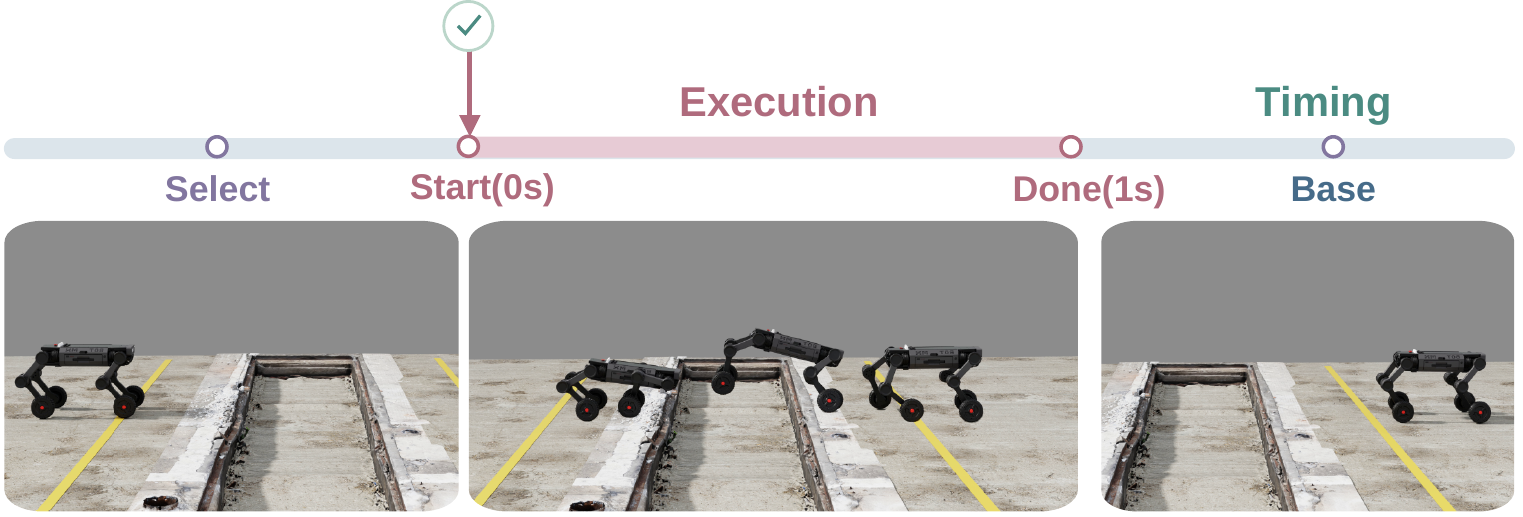}
\caption{\textbf{Gap maneuver illustrated in simulation.}
The sequence shows approach, gap crossing, and exit.
The reported success counts are obtained from separate
physical-robot trials.}
\label{fig:gap_sequence}
\end{figure}

\paragraph{Single-reference trials on the physical robot.}
We evaluated Guard and Gap on the physical robot, with ten trials
per skill from a prescribed approach to a designated exit.
The motor policies continued to receive live proprioception
throughout each maneuver (Sec.~\ref{sec:single_reference}).
Figures~\ref{fig:guard_sequence} and~\ref{fig:gap_sequence}
provide simulation illustrations of the maneuver sequences.
All success counts reported in this subsection were obtained
from physical-robot trials.

\paragraph{Guard.}
Guard retained selected localization-pose channels during execution
and successfully completed all ten physical-robot trials.
Figure~\ref{fig:guard_sequence} illustrates the skill's approach,
rodent-barrier crossing, and exit in simulation.

\paragraph{Gap.}
Gap retained its perceptual vector at takeoff and successfully
completed all ten physical-robot trials.
Figure~\ref{fig:gap_sequence} illustrates the corresponding
approach, gap crossing, and exit in simulation.

\paragraph{Results.}
Table~\ref{tab:reference_trials} summarizes the physical-robot
results, with both Guard and Gap achieving $10/10$ successful
trials. These results demonstrate the feasibility of executing
both skills on the physical robot with their respective
input-retention rules under the tested obstacle and approach
conditions.

\begin{table}[!htbp]
\centering\footnotesize
\setlength{\tabcolsep}{3pt}
\caption{\textbf{Single-reference robot trials.} A successful trial
crosses the obstacle and reaches the designated exit.}
\label{tab:reference_trials}
\begin{tabularx}{\linewidth}{@{}lXc@{}}
\toprule
Skill & Retained input & Successes ($n/10$)\\
\midrule
Guard & Selected localization pose & $10/10$\\
Gap & Perceptual vector at takeoff & $10/10$\\
\bottomrule
\end{tabularx}
\end{table}

\paragraph{Model choice within SkillHarness.}
We use Gemma~4 E2B for local visual monitoring and DeepSeek
Flash for cloud planning. The local model choice was informed
by a preliminary replay study on 18 labelled development images.
For single-pass skill classification, Gemma achieved 14/18
correct predictions (77.8\%) with a median latency of 1.289\,s.
The best tested Qwen3.5-0.8B configuration achieved 8/18
(44.4\%) at 1.035\,s and misclassified all seven
\textsc{Climb} examples. We therefore retained Gemma despite
its modest latency overhead. These results were obtained
with model-specific prompting and serve as development
evidence for model selection; they do not measure
end-to-end route completion.

\paragraph{Cloud-planning cost.}
We use DeepSeek Flash in non-thinking mode for repeated visual
planning requests. Using the arithmetic mean of its quoted API
rates, a request with 10,000 uncached input tokens and 100
billable output tokens has an estimated cost of US\$0.00234.
For the same billed token budget, GPT-6.1 Sol costs US\$0.021,
corresponding to an estimated cost reduction of 88.9\% with
DeepSeek Flash.

\section{Future Work}
\label{sec:discussion}

Future work will investigate compositional generalization across
broader skill repertoires, task scenarios, and robot platforms.
We will explore how skill contracts can represent capability
uncertainty and transition requirements, allowing the same
coordination framework to organize heterogeneous policies for
unseen task combinations. Extending this formulation to tasks
that combine locomotion, inspection, and interaction will further
test the generality of execution-aware coordination.

Another direction is to learn reusable execution constraints
from experience. We will investigate how logged outcomes and
failure cases can inform revisions to Harness admission rules
while keeping the motor policies fixed. A central question is
whether rules acquired from a limited set of interactions can
generalize to unseen obstacle arrangements and task sequences.
Offline replay and held-out evaluation will support the study
of rule transfer and the preservation of previously acquired
capabilities.

We will also investigate the interplay between reasoning latency
and closed-loop task reliability. For a moving robot, the value
of a decision depends on both its semantic correctness and its
continued applicability when inference completes. Adaptive
reasoning mechanisms could allocate computational effort according
to scene uncertainty, task urgency, and execution phase.
Jointly considering when to trigger reasoning, how much computation
to allocate, and when to admit an action may improve the
framework's response speed while maintaining reliable long-horizon
skill coordination.

Finally, we will extend the framework to dynamic environments
and interrupted tasks. This direction requires studying how
task-level replanning, recovery behaviors, and phase-dependent
reference updates can be coordinated under uncertain execution
progress. The goal is to adapt local decisions to environmental
changes while preserving progress toward the persistent task
objective.

\section{Conclusion}
SkillHarness coordinates independent locomotion policies through
event-triggered visual reasoning and execution-time proposal
verification. Local geometry, task context, and phase permissions
determine when a proposed goal or skill can take effect. Each
policy receives inputs through its own adapter, which specifies
how obstacle measurements and body feedback are used during
traversal. On the physical robot, Guard and Gap each completed
ten out of ten single-reference trials under the tested conditions.
The route protocol evaluates end-to-end completion as terrain
composition and geometry vary, and the matched verification
ablation tests the contribution of the Harness checks to successful
execution and skill switching.

\section*{LLM Usage Statement}
We used OpenAI ChatGPT to assist with English language polishing
and improve the clarity of the manuscript, and OpenAI Codex to
assist with code development. The authors take full responsibility
for the final manuscript, code, and reported results.

{\small
\bibliographystyle{ieeenat_fullname}
\bibliography{references}
}
\end{document}